%% file: preprint.tex
\documentclass{article}

\usepackage[preprint]{neurips_2025}

\usepackage[utf8]{inputenc} 
\usepackage[T1]{fontenc}    
\usepackage{hyperref}       
\usepackage{url}            
\usepackage{booktabs}       
\usepackage{amsfonts}       
\usepackage{nicefrac}       
\usepackage{microtype}      
\usepackage{xcolor}         
\usepackage{amsmath}
\usepackage{multirow}
\usepackage{cite}
\usepackage{natbib}
\usepackage{graphicx}
\usepackage{array}
\usepackage{multirow}
\usepackage{multicol}
\usepackage{booktabs}
\usepackage{color}
\usepackage{tcolorbox}
\usepackage{verbatim}
\usepackage{colortbl} 
\usepackage{vcell}
\usepackage{arydshln}
\usepackage{amsmath}
\usepackage{amsfonts}
\usepackage{tabularx}
\usepackage{makecell}
\usepackage[normalem]{ulem}
\usepackage{CJKutf8}
\usepackage{pifont}
\usepackage{tikz}
\usepackage{pgfplots}
\usepackage{inconsolata}
\usepackage{fancyvrb}
\usepackage{todonotes}
\usepackage{listings}
\usepackage{pifont}

\newcommand{\chinese}[1]{{\begin{CJK*}{UTF8}{gkai} #1 \end{CJK*}}}
\newcommand{\cmark}{{\textbf{\textcolor[rgb]{0.1, 0.5, 0.1}{\ding{51}}}}}
\newcommand{\xmark}{{\textbf{\color{red}{\ding{55}}}}}

\title{MT$^{3}$: Scaling MLLM-based Text Image Machine Translation via Multi-Task Reinforcement Learning}

\makeatletter
\def\thanks#1{\protected@xdef\@thanks{\@thanks
        \protect\footnotetext{#1}}}
\makeatother
\author{
    Zhaopeng Feng$^{1\ast}$\quad
    Yupu Liang$^{2\ast}$\quad
    Shaosheng Cao$^{3}$\textsuperscript{\ding{41}} \quad
    Jiayuan Su$^{1}$  \quad 
    Jiahan Ren$^{1}$ \quad \\
    \bf Zhe Xu$^{3}$ \quad 
    \bf Yao Hu$^{3}$ \quad
    \bf Wenxuan Huang$^{4}$ \quad 
    \bf Jian Wu$^{1}$ \quad
    \bf Zuozhu Liu$^{1}$\textsuperscript{\ding{41}} \\
    $^{1}$Zhejiang University \quad
    $^{2}$University of Chinese Academy of Sciences \quad \\
    $^{3}$Xiaohongshu Inc. \quad 
    $^{4}$East China Normal University \quad \\
    \texttt{\{zhaopeng.23,zuozhuliu\}@intl.zju.edu.cn, }\\
    \texttt{liangyupu2021@ia.ac.cn} \\
    \texttt{\{caoshaosheng,qiete,xiahou\}@xiaohongshu.com} \\
}

\thanks{$^{\ast}$ \space Equal Contribution.}
\thanks{\textsuperscript{\ding{41}} \space Corresponding author.}

\begin{document}

\maketitle

\begin{abstract}

Text Image Machine Translation (TIMT)—the task of translating textual content embedded in images—is critical for applications in accessibility, cross-lingual information access, and real-world document understanding. However, TIMT remains a complex challenge due to the need for accurate optical character recognition (OCR), robust visual-text reasoning, and high-quality translation, often requiring cascading multi-stage pipelines. Recent advances in large-scale Reinforcement Learning (RL) have improved reasoning in Large Language Models (LLMs) and Multimodal LLMs (MLLMs), but their application to end-to-end TIMT is still underexplored. To bridge this gap, we introduce \textbf{MT$^{3}$}, the first framework to apply \textbf{M}ulti-\textbf{T}ask RL to \textbf{M}LLMs for end-to-end \textbf{T}I\textbf{MT}. \textbf{MT$^{3}$} adopts a multi-task optimization paradigm targeting three key sub-skills: text recognition, context-aware reasoning, and translation. It is trained using a novel multi-mixed reward mechanism that adapts rule-based RL strategies to TIMT’s intricacies, offering fine-grained, non-binary feedback across tasks.  Furthermore, to facilitate the evaluation of TIMT in authentic cross-cultural and real-world social media contexts, we introduced XHSPost, the first social media TIMT benchmark. Our \textbf{MT$^{3}$}-7B-Zero achieves state-of-the-art results on the latest in-domain MIT-10M benchmark, outperforming strong baselines such as Qwen2.5-VL-72B and InternVL2.5-78B by notable margins across multiple metrics. Additionally, the model shows strong generalization to out-of-distribution language pairs and datasets. In-depth analyses reveal how multi-task synergy, reinforcement learning initialization, curriculum design, and reward formulation contribute to advancing MLLM-driven TIMT. 

\end{abstract}

\section{Introduction}
Text Image Machine Translation (TIMT) is a crucial subfield of machine translation (MT) that focuses on translating source-language texts embedded in images into target-language texts~\citep{ma2022improving,lan2023exploring,liang2024document}. It has been widely applied in scenarios including photo translation, scanned document translation, and screenshot translation. Unlike traditional text-based machine translation~\citep{xu2024xalma, feng2024improving}, where input and output are purely textual, TIMT is inherently a cross-modal task. It requires systems to process images and generate corresponding textual translations, necessitating the comprehensive integration of fine-grained textual content and diverse visual elements such as layout, objects, and color schemes to produce accurate translations. However, current TIMT approaches, both cascade systems~\citep{hinami2021towards, sable2023doc, zhang2023novel, zhang2025understand} and end-to-end (E2E) models~\citep{zhu2023peit, lan2023exploring, ma2024born,  liang2024document}, often do not explicitly capture fine-grained textual information or model the integration of visual element comprehension. Cascade systems typically use OCR output for a text-based MT model, thereby overlooking visual information, and suffer from error propagation. E2E models, while aiming for unified training, often lack explicit modeling of OCR and visual understanding, which is beneficial for translation quality~\citep{niu2024umtit}.

Recent advances in Multimodal Large Language Models (MLLMs) have demonstrated strong capabilities in capturing fine-grained textual information and understanding non-textual content across various cross-modal tasks, such as OCR and VQA~\citep{qwen2_5vl, internvl, qvq-72b-preview}. Concurrently, the efficacy of large-scale Reinforcement Learning (RL) in substantially enhancing the reasoning capabilities of Large Language Models (LLMs)~\citep{guo2025deepseek, kimi, qwq32b} has driven its adoption to MLLMs. Several studies have successfully adapted R1-like RL paradigms, often using synthesized reasoning data or rule-based visual rewards, to enhance MLLM capabilities in tasks like multimodal mathematical reasoning or visual understanding~\citep{huang2025vision, zhang2025r1, shen2025vlm, liu2025visualrft, meng2025mmeureka, chen2025r1v}. Given that RL training can enable models to rapidly learn specific output and reasoning patterns, these concurrent developments motivate our exploration into whether such R1-like training paradigms can be effectively applied to scale up the TIMT capabilities of MLLMs.

In this work, we introduce \textbf{MT$^{3}$}, the first framework to apply \textbf{M}ulti-\textbf{T}ask RL to \textbf{M}LLMs for end-to-end \textbf{T}ext \textbf{I}mage \textbf{M}achine \textbf{T}ranslation. \textbf{MT$^{3}$} adopts a multi-task optimization paradigm targeting three key sub-skills: text recognition, context-aware reasoning, and translation. It is trained using a novel multi-mixed reward mechanism that adapts rule-based RL strategies to TIMT’s intricacies, offering fine-grained feedback across tasks. Furthermore, recognizing the need for evaluations that reflect real-world social media interactions, we introduce XHSPost, the first social media TIMT benchmark, to evaluate TIMT in more authentic cross-cultural scenarios. Our experiments demonstrate the efficacy of this approach: our \textbf{MT$^{3}$}-7B-Zero model achieves state-of-the-art results on the latest in-domain MIT-10M benchmark~\citep{li2024mit10m}, outperforming strong MLLM baselines, including Qwen2.5-VL-72B~\citep{qwen2_5vl} and InternVL2.5-78B~\citep{internvl}, by notable margins (approximately 15-25 points average improvement across BLEU, chrF++, METEOR). Additionally, the model shows strong generalization to out-of-distribution (OOD) language pairs and datasets, including the newly introduced XHSPost. In-depth analyses reveal how multi-task synergy, reinforcement learning initialization, curriculum design, and reward formulation contribute to advancing MLLM-driven TIMT. Our core contributions are as follows:
\begin{itemize}

\item We propose MT$^{3}$, the first framework applying multi-task reinforcement learning to MLLMs for end-to-end TIMT, featuring a novel multi-task optimization paradigm and an effective multi-mixed reward mechanism. 
Extensive experiments validate MT$^{3}$'s superior performance on standard TIMT benchmarks and its strong OOD generalization capability, significantly surpassing existing cascaded systems and advanced MLLMs.

\item We present XHSPost, the first social media TIMT benchmark. This new resource is designed to facilitate research and enable practical evaluation of TIMT systems in authentic, real-world cross-cultural communication scenarios.

\item We provide comprehensive analyses yielding key insights into R1-like RL for TIMT. We demonstrate that multi-task synergy is vital for R1-like RL success in TIMT, and zero-start RL from MLLMs outperforms SFT+long Chain-of-Thought (CoT) in performance and efficiency. Besides, we show that curriculum learning and reward metric selection critically impact training and results.

\end{itemize}

\section{Related Works}
\label{sec:related_works}

\noindent{\textbf{Text Image Machine Translation.}}
Text Image Machine Translation (TIMT) aims to translate texts embedded in images \citep{ma2022improving}. Prevailing approaches include:
(1) \textit{Cascade systems}~\citep{hinami2021towards, sable2023doc, zhang2023novel, zhang2025understand}, which sequentially combine Optical Character Recognition (OCR) and Neural Machine Translation (NMT), often facing issues like error propagation and latency.
(2) \textit{End-to-end (E2E) models}~\citep{zhu2023peit, lan2023exploring, ma2024born, niu2024umtit, liang2024document}, developed to unify training and improve efficiency. Early E2E methods integrated visual encoders and text decoders, with some bridging modality gaps using pre-trained components~\citep{zhu2023peit} or dynamically assembling models~\citep{liang2024document}. While advanced MLLMs~\citep{internvl, qwen2_5vl} show promise for more effective E2E TIMT~\citep{li2024mit10m}, their specific application to this task remains underexplored.

\noindent{\textbf{MLLM Reasoning with Reinforcement Learning.}}
RL has been shown to enhance the reasoning capabilities of LLMs~\citep{team2025kimi, guo2025deepseek}, leading to explorations of RL in MLLMs~\citep{huang2025vision, zhang2025r1, shen2025vlm, liu2025visualrft,meng2025mmeureka}.
Several studies employ R1-like training methodologies, often cultivating a "\verb|<think>| then \verb|<answer>|" paradigm. For instance, some focus on improving MLLM reasoning through synthesized data~\citep{huang2025vision}, while others design online RL frameworks for self-improvement via step-wise rewarding~\citep{zhang2025r1}, or explore R1-style RL for general vision-language tasks~\citep{shen2025vlm}.
These efforts have primarily concentrated on tasks like multimodal mathematical reasoning or general visual understanding. However, there has been less focus on leveraging multi-task RL paradigms or addressing application-oriented downstream tasks like TIMT, a gap our work aims to fill.

\section{Method}
\label{sec:method}
In this section, we detail the MT$^{3}$ framework. Our MT$^{3}$ framework decomposes TIMT into distinct, explicit sub-tasks and employs a novel multi-mixed reward mechanism to guide the MLLM. The training is performed using the Group Relative Policy Optimization (GRPO)~\citep{shao2024deepseekmath} algorithm, selected for its efficiency in training LLMs with RL.

\subsection{Multi-Task Formulation for TIMT}
\label{sec:method_multitask_formulation}
TIMT is intrinsically linked to fundamental MLLM capabilities like visual perception (specifically, OCR) and multimodal reasoning. We hypothesize that enhancing these core capabilities while simultaneously optimizing for translation quality will synergistically boost the MLLM's overall TIMT proficiency. To this end, MT$^{3}$ formulates TIMT as a multi-task learning problem where the MLLM performs a sequence of interconnected tasks within a unified generation process. This process is guided by the structured prompt, which instructs the model to generate its output in a specific format encompassing recognition, reasoning, and translation.
\vspace{-1.5mm}
\begin{tcolorbox}[
    colframe=teal!70!black, 
    colback=teal!10!white, 
    coltitle=white, 
    fonttitle=\bfseries, 
    title=MT$^{3}$ Prompt, 
    sharp corners, 
    boxrule=0.5mm, 
]
System: You are a helpful translation assistant. The user provides an image containing \textit{\{source\_language\}} text and asks for the corresponding \textit{\{target\_language\}} translation. First, the assistant recognizes all the text in the image following the natural reading order. Then, the assistant carefully analyzes the recognized text and the visual elements in the image, considering the layout, objects, color schemes, spatial relationships, and other contextual clues that may influence meaning. This integrated understanding ensures the translation is accurate, coherent, and appropriate to the visual setting. After thorough reasoning based on both textual content and visual context, the assistant provides the user with the final translation in reading order. The recognized text, reasoning process, and final translation are enclosed within <recognize> </recognize>, <think> </think>, and <translate> </translate> tags, respectively. The format must be as follows: <recognize> recognized text here </recognize> <think> reasoning process here </think> <translate> final translation here </translate> \\
User: \textit{\{image\}} Translate all the text in this image into \textit{\{target\_language\}} following the natural reading order.
\label{prompt:full_task}
\end{tcolorbox}
\vspace{-1.5mm}
Here, \textit{\{source\_language\}} and \textit{\{target\_language\}} indicate the source and target languages, and \textit{\{image\}} denotes the input image requiring translation.

\noindent{\textbf{Recognition.}} The MLLM first performs OCR, transcribing image text in its natural reading order within \verb|<recognize>| and \verb|</recognize>| tags. This step directly leverages the model's visual perception and text extraction capabilities.

\noindent{\textbf{Reasoning.}} Subsequently, the MLLM engages in a reasoning process, outputting its analysis within \verb|<think>| and \verb|</think>| tags. It considers the recognized text alongside the broader visual context (e.g., layout, objects, color schemes, spatial relationships) to ensure the subsequent translation is contextually appropriate. This task explicitly cultivates the MLLM's multimodal reasoning for TIMT.

\noindent{\textbf{Translation.}} Finally, informed by the recognition and reasoning stages, the MLLM generates the target translation of the recognized text. This output is presented within \verb|<translate>| and \verb|</translate>| tags, maintaining the reading order and focusing on context-enhanced translation.

This explicit multi-task formulation guides the model through distinct recognition, reasoning, and translation stages. It allows for monitoring and rewarding these intermediate steps, thereby fostering more robust and interpretable TIMT capabilities, as validated in our ablation studies (Section~\ref{sec:analysis_multitask_ablation}).

\subsection{Multi-Mixed Reward Mechanism}
\label{sec:method_reward}
The reward signal $r$ is crucial in RL. While rule-based rewards suit tasks with verifiable answers (e.g., math, coding)~\citep{guo2025deepseek}, translation often lacks a single correct output and TIMT's multi-faceted quality (e.g., recognition accuracy, translation quality) necessitates a more nuanced approach. We propose a \textbf{multi-mixed reward mechanism}, adapting rule-based concepts by integrating format adherence checks with task-specific quality assessments.

\noindent{\textbf{Format Reward ($R_{format}$).}} We use regular expression extraction to enforce the structured response format illustrated in the MT$^{3}$ prompt. The format reward is computed as:
$$ R_{format} = \begin{cases} 1, & \text{if format is correct} \\ -3, & \text{if format is incorrect} \end{cases} $$
This penalty for incorrect format strongly discourages deviations and encourages the model to rapidly learn the required output structure, as evidenced in Figure~\ref{fig:task_ablation}.

\noindent{\textbf{Task-Specific Rewards ($R_{task}$).}} If the output format is correct ($R_{format}=1$), we calculate rewards for recognition and translation sub-tasks. While the MT community has developed various evaluation metrics~\citep{freitag-etal-2022-results, freitag-etal-2023-results}, \citet{feng2025mtr1zero} and \citet{ramos2025finegrained} noted that relying on a single metric might result in sub-optimal overall performance. Therefore, we propose a metric-mixed approach. We average multiple standard metrics for each task, providing a more robust, balanced, and nuanced reward signal:

\begin{itemize}
    \item \textbf{Translation Reward ($R_{task-trans}$):} Assesses final translation quality by averaging scores from standard MT metrics: BLEU~\citep{post-2018-call}, chrF++~\citep{popovic2017chrf++}, and METEOR~\citep{banerjee2005meteor}, against the reference. Formally, $R_{task-trans} = (S_{\text{BLEU}} + S_{\text{METEOR}} + S_{\text{chrF++}}) / 3$. The specific combination of metrics can be varied, as explored in our experiments on metric reward selection (Section~\ref{sec:analysis_metric_reward}). Furthermore, as highlighted by \citet{feng2025mtr1zero}, the quality of the reasoning process is implicitly encouraged through its positive influence on the final translation quality.
    
    \item \textbf{Recognition Reward ($R_{task-rec}$):} Evaluates recognized text quality against a ground-truth transcription by averaging suitable OCR metrics~\citep{fu2024ocrbenchv2}: BLEU, METEOR, F1-score, and normalized versions of Character Error Rate (CER) and Edit Distance. Formally, $R_{task-rec} = (S_{\text{BLEU}} + S_{\text{METEOR}} + S_{\text{F1-score}} + \min(\max(1-S_{\text{EditDistance}}, 0), 1) + \min(\max(1-S_{\text{CER}}, 0), 1)) / 5$.
\end{itemize}

\noindent{\textbf{Final Reward.}} The final reward $r$ fed to GRPO combines $R_{format}$ and $R_{task}$:
$$ r = \begin{cases} R_{format} + R_{task-rec} + R_{task-trans}, & \text{if format is correct} \\ -3, & \text{if format is incorrect} \end{cases} $$
This formulation extends the simple binary rule-based reward (a base score of 1 for correct format) by adding continuous, mixed-metric scores for sub-tasks. This provides more granular feedback than purely binary rewards, enabling the model to learn fine-grained improvements in both recognition and translation quality, as evidenced by the training curves in Figure~\ref{fig:task_ablation}.

\subsection{RL Algorithm}
We use the GRPO algorithm~\citep{shao2024deepseekmath} to train the translation model with our rule-metric mixed reward. In each training step, for a given translational question $q$, we sample a group of candidate outputs $\{o_1,o_2,\cdots,o_G\}$ from the policy model $\pi_{\theta_{old}}$. $A_i = \frac{r_i - \operatorname{mean}(\{r_1, r_2, \dots, r_G\})}{\operatorname{std}(\{r_1, r_2, \dots, r_G\})}$ is the computed advantage using the group rule-metric mixed rewards $\{r_1,r_2,\cdots,r_G\}$. GRPO then maximizes the following objective function to optimize $\pi_{\theta}$:  
\begin{equation}
\resizebox{\textwidth}{!}{%
$J_{\mathrm{GRPO}}(\theta) = \mathbb{E}_{q \sim P(Q),\, \{o_i\}_{i=1}^G \sim \pi_{\theta_{\mathrm{old}}}(O \mid q)} \left[ \frac{1}{G} \sum_{i=1}^G \min\left( \frac{\pi_{\theta}(o_i \mid q)}{\pi_{\theta_{\mathrm{old}}}(o_i \mid q)} A_i,\, \mathrm{clip}\left( \frac{\pi_{\theta}(o_i \mid q)}{\pi_{\theta_{\mathrm{old}}}(o_i \mid q)},\, 1-\varepsilon,\, 1+\varepsilon \right) A_i \right) - \beta D_{\mathrm{KL}}\left( \pi_{\theta} \,\big\|\, \pi_{\mathrm{ref}} \right) \right]$
}
\label{eq1}
\end{equation}
where $\varepsilon$ and $\beta$ are hyperparameters controlling the PPO clipping threshold and the weight of the Kullback–Leibler (KL) divergence penalty~\citep{schulman2017proximal,shao2024deepseekmath}, respectively.

\subsection{Experimental Setup}
\label{sec:experimental_setup} 
\noindent{\textbf{Datasets and Evaluation Metrics.}}
Our primary experiments focus on English-Chinese (EN-ZH) and Chinese-English (ZH-EN) TIMT tasks. For RL training, we select 15K image-text pairs from the MIT-10M dataset~\citep{li2024mit10m} for each translation direction. To comprehensively evaluate our MT$^{3}$ framework, we test its performance across three distinct settings: 
(1) \textbf{In-domain (IND):} utilizing the standard ZH-EN and EN-ZH test sets from the MIT-10M dataset. 
(2) \textbf{Out-of-distribution (OOD) - Language Pairs:} employing unseen language pairs (EN-DE, ZH-FR, DE-FR) from within the MIT-10M dataset. 
(3) \textbf{Out-of-distribution (OOD) - Datasets:} using the test sets from the OCRMT30K dataset~\citep{lan2023exploring} and the document-level DoTA dataset~\citep{liang2024document}. 
For all evaluations, we report SacreBLEU~\citep{post-2018-call}, chrF++~\citep{popovic2017chrf++}, and METEOR~\citep{banerjee2005meteor} scores as previous works~\citep{li2024mit10m,liang2024document}.


\input{table/main}
\input{table/ood_chrf_meteor}

\noindent{\textbf{XHSPost Benchmark: A Real-World Social Media TIMT Scenario.}}
Existing TIMT datasets primarily focus on general photos, academic documents, or advertisements (see Figure~\ref{fig:dataset_case}). To address the notable absence of benchmarks for the increasingly prevalent social media domain, we constructed XHSPost, a dataset of social media posts for ZH-EN and EN-ZH TIMT using content from Xiaohongshu. After a meticulous selection and filtering process (detailed in Appendix~\ref{app:xhspost_construction}), we obtained 106 English and 109 Chinese posts. We anticipate XHSPost will foster real-world applications of TIMT.

\noindent{\textbf{Baselines.}}
To comprehensively assess the performance of our MT$^{3}$-7B-Zero, we compare it against a diverse set of baselines. This includes a \textbf{Cascade System}, which first employs EasyOCR\footnote{\url{https://github.com/JaidedAI/EasyOCR}} to extract text from images, arranges the extracted text sequentially, and then translates it using the NLLB-3.3B model~\citep{costa2022no}. We also benchmark against several \textbf{Advanced MLLMs} used in a zero-shot prompting setup, specifically the InternVL2.5 series~\citep{internvl} and Qwen2.5-VL series~\citep{qwen2_5vl}; more evaluation details can be found in Appendix~\ref{app:eval}. Furthermore, we compare against \textbf{Supervised Fine-Tuning (SFT)} variants of Qwen2.5-VL-7B. These SFT models were trained on the MIT-10M dataset using two distinct prompt templates: one focused solely on the end-to-end \textit{TIMT} task, and another employing an \textit{OCR + TIMT} multi-task format. The specific SFT prompts are also provided in Appendix~\ref{app:sft_prompts}.

\noindent{\textbf{Training Details.}}
Our implementation is based on verl\footnote{\url{https://github.com/volcengine/verl}} framework. We select the Qwen2.5-VL-7B as starting models for MT$^{3}$ training.
During training, we configure a batch size of 64 and utilize 16 rollouts per prompt within the GRPO algorithm. More training details are in Appendix~\ref{app:train_details}.

\subsection{Experimental Results}
\label{sec:main_result}

\noindent{\textbf{In-Domain Performance.}}
As detailed in Table~\ref{tab:main_mt_results}, our MT$^{3}$-7B-Zero significantly outperforms compared systems on the in-domain MIT-10M benchmark. For ZH-EN, MT$^{3}$-7B-Zero achieved an average score of 36.24, surpassing Qwen2.5-VL-72B (Avg. 28.47) by 7.77 points and InternVL2.5-78B by 14.09 points. The improvements are more pronounced for EN-ZH, where our model (Avg. 61.50) outperforms Qwen2.5-VL-72B by 15.14 points and InternVL2.5-78B by 17.46 points. Notably, our RL-trained model also surpasses SFT variants of Qwen2.5-VL-7B, including those trained with OCR+TIMT multi-task or single TIMT task setups. These results underscore the efficacy of our multi-task RL approach over supervised fine-tuning and zero-shot larger MLLMs for TIMT.

\noindent{\textbf{Out-of-Distribution Performance.}}
We assessed MT$^{3}$-7B-Zero's generalization ability on diverse out-of-distribution (OOD) scenarios, including unseen language pairs from MIT-10M (EN-DE, ZH-FR, DE-FR), distinct TIMT datasets (OCRMT30K~\citep{lan2023exploring}, the document-level DoTA~\citep{liang2024document}), and our newly introduced real-world social media benchmark, XHSPost. As shown in Table~\ref{tab:ood_chrf_meteor}, our model consistently demonstrated strong OOD performance, generally outperforming comparable-sized models and rivaling larger MLLMs. For instance, on MIT-10M (DE-FR), MT$^{3}$-7B-Zero achieved 54.99 chrF++ and 53.88 METEOR, surpassing all listed baselines. On DoTA, our model's 46.60 chrF++ was considerably higher than Qwen2.5-VL-72B's 37.65, with a competitive METEOR score. These findings highlight the robust generalization fostered by the MT$^{3}$ framework. Appendix~\ref{app:ood_bleu} provides corresponding OOD BLEU scores.

\input{table/task_ablation}

\begin{figure}[t]
\centering
\includegraphics[width=0.9\textwidth]{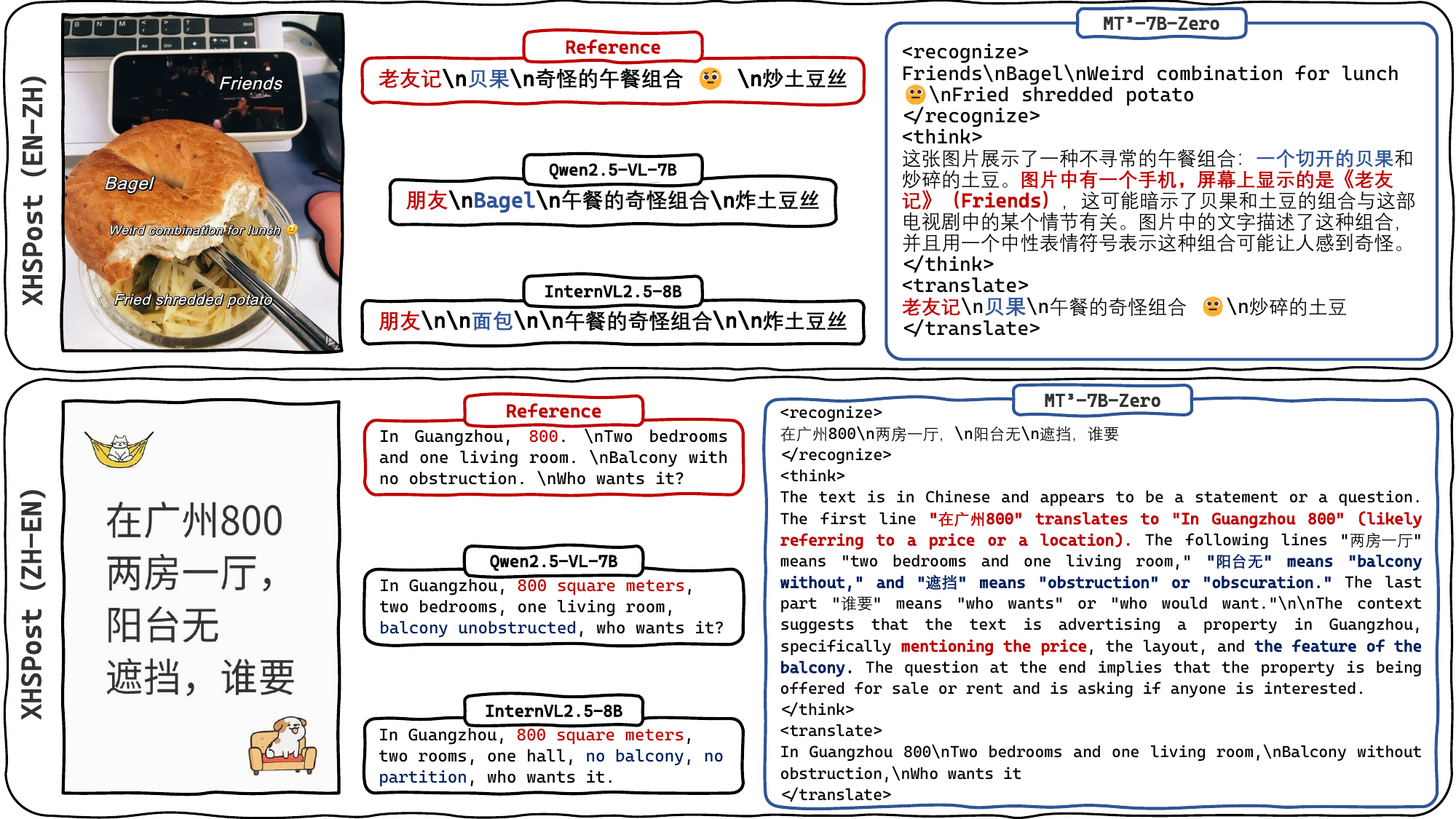}
\caption{Illustrative TIMT examples from the XHSPost benchmark. MT$^{3}$ demonstrates superior contextual understanding by correctly translating '\textit{Friends}' to '\textit{\chinese{老友记}}' based on visual context (\textbf{Top}), and accurately interpreting a property advertisement, including price and layout details (\textbf{Bottom}).
}
\label{fig:xhs_case}
\vspace{-3mm}
\end{figure}

\section{Analyses}
\label{sec:analyses}

\subsection{Multi-Task Ablation Analysis}
\label{sec:analysis_multitask_ablation}
\begin{figure}[t]
\centering
\includegraphics[width=0.9\textwidth]{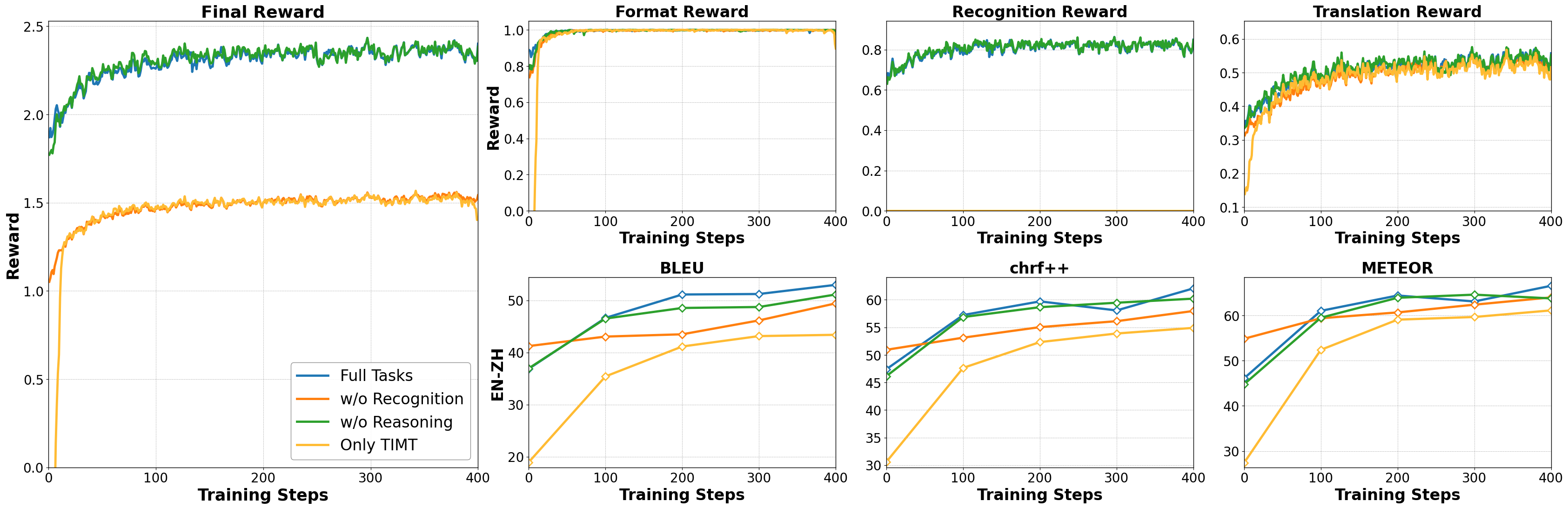}
\caption{Impact of multi-task ablation on reward optimization and performance progression. \textbf{Top row}: Progression of \textit{Final Reward} and individual task rewards (Format, Recognition, Translation). \textbf{Bottom row}: Progression of translation quality metrics (BLEU, chrF++, METEOR).}
\label{fig:task_ablation}
\vspace{-3mm}
\end{figure}

To disentangle the contributions of the text recognition, reasoning, and translation components within our multi-task framework, we conduct an ablation study comparing the full-task MT$^{3}$-7B-Zero against variants where specific tasks are omitted during RL training: (1) \textit{w/o Reasoning}; (2) \textit{w/o Recognition}; and (3) \textit{Only TIMT}. Detailed prompts for these settings are available in Appendix~\ref{app:multi_task_prompt}. As shown in Table~\ref{tab:ablation_results}, the \textit{Full Tasks} setting consistently achieves the best performance across the evaluated benchmarks. Removing the explicit text recognition step (\textit{w/o Recognition}) leads to a significant performance drop across all settings, underscoring the criticality of accurate visual text extraction. Omitting the reasoning step (\textit{w/o Reasoning}) generally results in a performance decrease, particularly on more nuanced datasets, although its utility can vary with task complexity or language pair characteristics. Training only the translation task (\textit{Only TIMT}) results in the most substantial degradation. As illustrated in Figure~\ref{fig:xhs_case}, MT$^{3}$ correctly translates culturally nuanced terms and interprets ambiguous numerical information by integrating visual and textual cues, demonstrating how explicit recognition and reasoning foster deeper contextual understanding crucial for TIMT(see Appendix~\ref{app:xhs_case_analysis} for more detailed analysis). These cases demonstrate how the explicit modeling of recognition and reasoning within MT$^{3}$ fosters deeper contextual understanding crucial for accurate TIMT in diverse scenarios. The reward and performance progression curves (Figure~\ref{fig:task_ablation}) further support this: variants with recognition yield higher translation rewards, and the \textit{Full Tasks} setting shows the most effective trajectories. This confirms the benefit of our multi-task formulation, emphasizing explicit text recognition and the synergy of jointly optimizing all three sub-tasks.

\input{table/think_ablation}

\subsection{Impact of Initialization Strategy: Zero vs. QVQ-Distill}
\label{sec:analysis_init_strategy}
To assess the effect of the RL starting point, we compare MT$^{3}$-7B-Zero ( RL directly from the general MLLM checkpoint, \textit{Zero-start}) against MT$^{3}$-7B-QVQ-Distill (initialized via SFT on 10K high-quality TIMT examples distilled from the QVQ-72B~\citep{qvq-72b-preview}; details in Appendix~\ref{app:initial}). Training dynamics (Figure~\ref{fig:qvq_zero}) show that while the QVQ-Distill model benefits from an initial SFT performance boost, the Zero-start model exhibits a steeper learning curve, rapidly surpassing the cold-start variant and converging to a markedly higher performance ceiling. The Zero-start model also produces considerably shorter and more stable response lengths. A qualitative example in Appendix~\ref{app:case} further reveals differences in reasoning patterns: the QVQ-Distill model often simulates self-reflection patterns, whereas the Zero-start model can incentivize a more straightforward and less redundant reasoning path conducive to TIMT. Table~\ref{tab:think_ablation} corroborates these findings, showing Zero-start yields significantly better final performance.

\begin{figure}[t]
\centering
\includegraphics[width=0.9\textwidth]{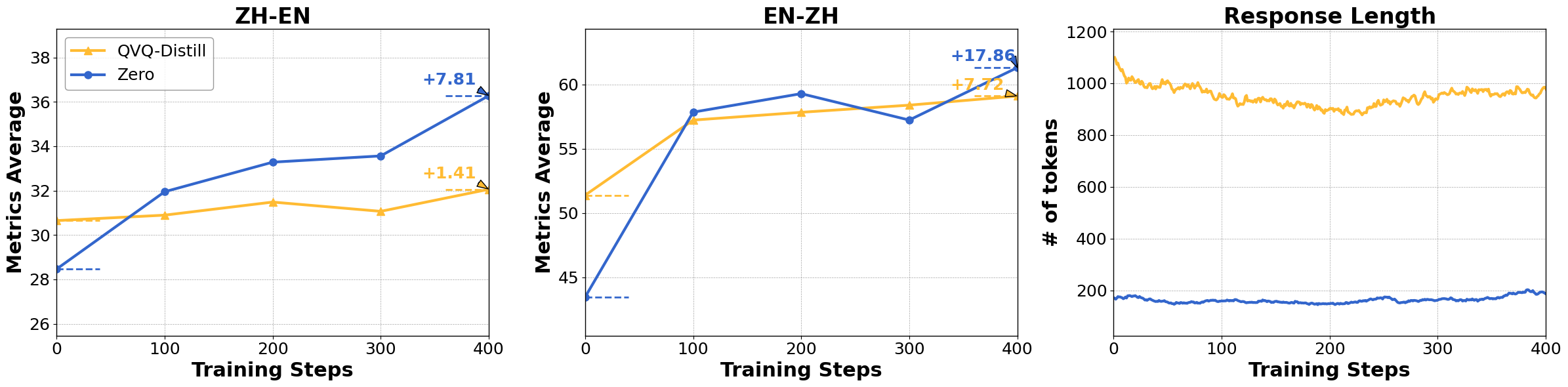}
\caption{
Training dynamics comparing Zero-start RL (MT$^{3}$-7B-Zero) vs. SFT initialization (MT$^{3}$-7B-QVQ-Distill). \textbf{Left and Center:} Average metric score progression on MIT-10M ZH-EN and EN-ZH test sets. \textbf{Right:} Average response length during RL training.
}
\label{fig:qvq_zero}
\vspace{-2mm}
\end{figure}

\subsection{Curriculum Learning}
\label{sec:analysis_curriculum} 
\begin{figure}[h]
\centering
\includegraphics[width=0.95\textwidth]{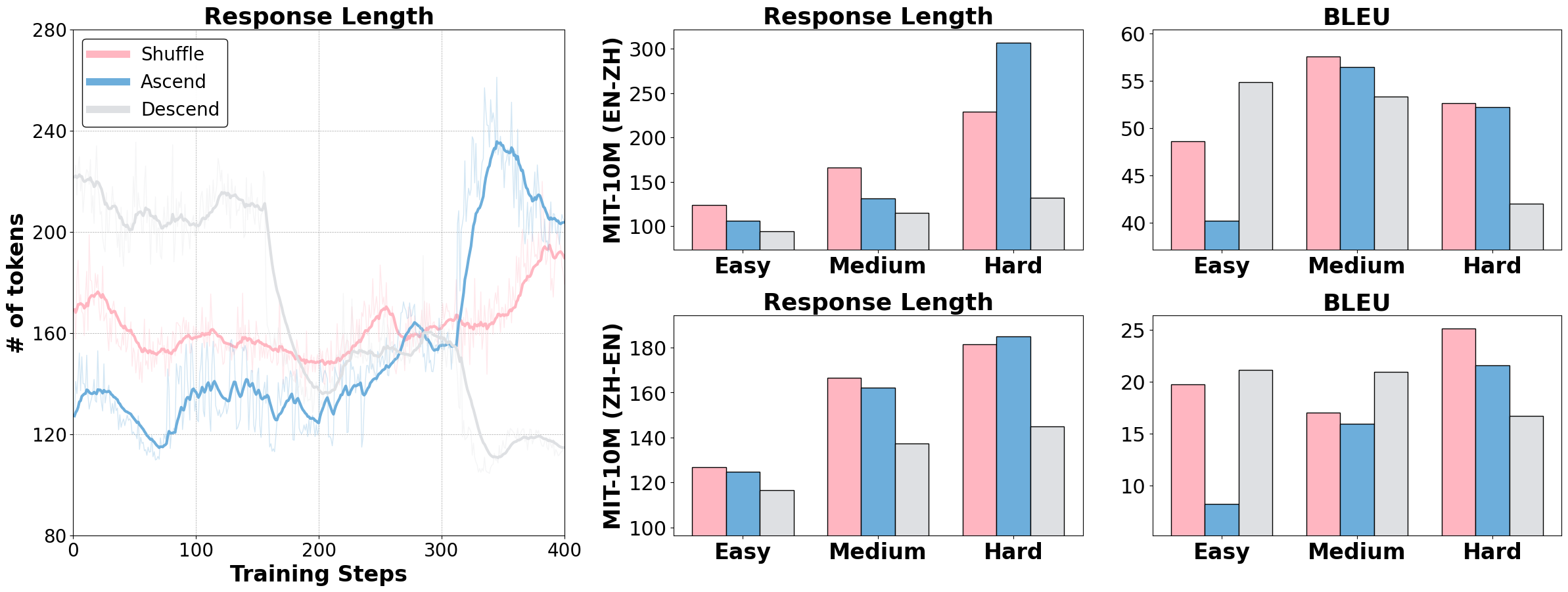}
\caption{Influence of curriculum learning strategies on training dynamics and performance. \textbf{Left}: Average response length dynamics for \textit{Shuffle}, \textit{Ascending (easy-to-hard)}, and \textit{Descending (hard-to-easy)} difficulty curricula during training. \textbf{Right}: Final response length and BLEU scores on MIT-10M difficulty splits (Easy, Medium, Hard) for EN-ZH and ZH-EN..
}
\label{fig:curriculum}
\vspace{-3mm}
\end{figure}

We explore curriculum learning by ordering training data by difficulty, categorized by~\citet{li2024mit10m} as Easy, Medium, or Hard using heuristics based on text length and scene complexity, with details in Appendix~\ref{app:difficulty}. We compare three data presentation strategies: \textit{Shuffle} (random), \textit{Ascend} (easy-to-hard), and \textit{Descend} (hard-to-easy). Figure~\ref{fig:curriculum} shows response length dynamics: \textit{Ascend} leads to longer responses as harder samples appear, while \textit{Descend} starts longer and shortens. Figure~\ref{fig:curriculum} also shows the final BLEU scores and response lengths on MIT-10M test splits. \textit{Ascend} tends to yield competitive or superior BLEU scores on Hard instances, whereas \textit{Descend} performs better on Easy instances. \textit{Shuffle} demonstrates the most balanced performance across different difficulty splits. These results suggest that while specific curricula can tune for specific difficulty levels, a shuffled curriculum may be more broadly beneficial for robust TIMT.

\subsection{Metric Reward Selection for Translation}
\label{sec:analysis_metric_reward}
\begin{figure}[t]
\centering
\includegraphics[width=0.9\textwidth]{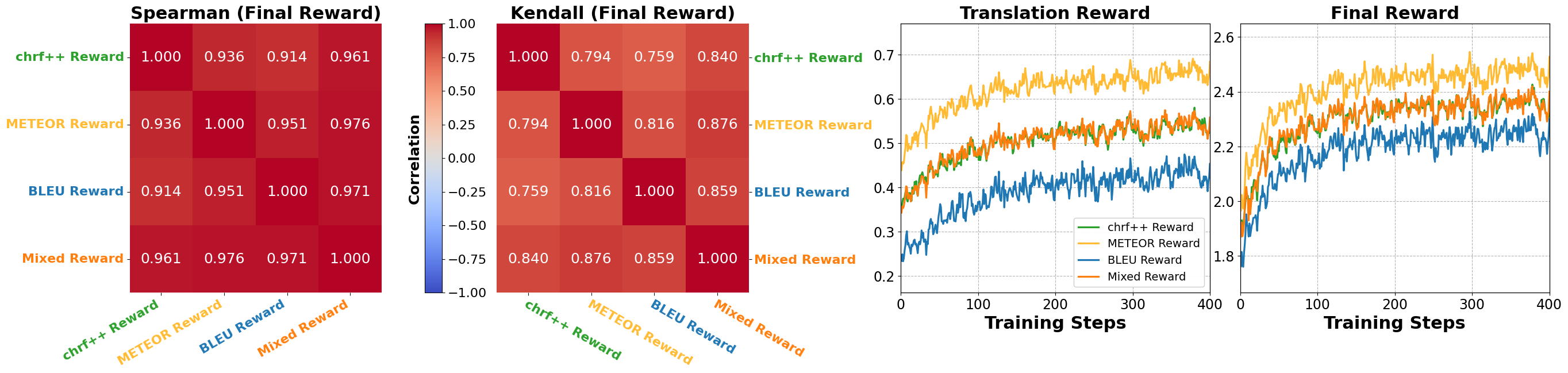}
\caption{
Analysis of translation rewards. \textbf{Left}: Kendall and Spearman correlation matrices between different individual metric rewards (chrF++, METEOR, BLEU) and the \textit{Mixed Reward}, based on final reward. \textbf{Right}: Progression of \textit{Translation Reward} and \textit{Final Reward} when optimizing for individual metric reward versus the \textit{Mixed Reward}.
}
\label{fig:metric_reward}
\vspace{-2mm}
\end{figure}

\begin{figure}[t]
\centering
\includegraphics[width=0.9\textwidth]{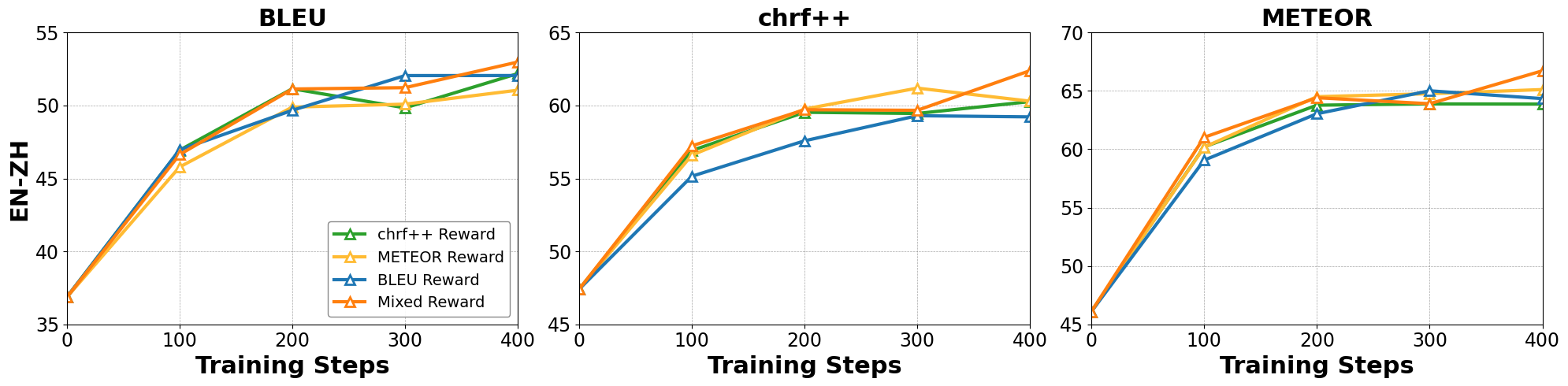}
\caption{ Effect of different translation reward metric choices ($R_{task-trans}$) on performance progression for EN-ZH translation on MIT-10M. Curves display BLEU, chrF++, and METEOR scores over training steps when optimizing with individual metrics versus the \textit{Mixed Reward}.
}
\label{fig:metric_metric}
\vspace{-2mm}
\end{figure}

Our multi-mixed reward mechanism (Section~\ref{sec:method_reward}) averages multiple metrics for task-specific rewards. Here, we analyze using individual metrics (BLEU, chrF++, METEOR) versus our proposed \textit{Mixed Reward} (average of these three) for $R_{task-trans}$. Figure~\ref{fig:metric_reward} presents Spearman and Kendall correlation matrices, which indicate positive correlations between the final rewards from individual metrics and the mixed-metric reward, suggesting the shared underlying signal of quality. Notably, BLEU shows the lowest correlation with chrF++, a divergence that is also reflected in their differing performance trajectories in Figure~\ref{fig:metric_metric}. These performance plots demonstrate that the \textit{Mixed Reward} generally results in the most consistent and often the highest performance across all three evaluation metrics. This mitigates optimizing for a single metric's peculiarities, leading to holistically improved translation quality and validating our mixed-metric approach.

\section{Conclusion and Future Work}
\label{sec:conclusion}
We introduce MT$^{3}$, the first framework to leverage multi-task reinforcement learning for end-to-end MLLM-based Text Image Machine Translation. By explicitly optimizing text recognition, context-aware reasoning, and translation through a novel multi-mixed reward mechanism, MT$^{3}$ significantly advances MLLM-based TIMT capabilities. Our experiments demonstrate that MT$^{3}$-7B-Zero achieves leading performance on the MIT-10M benchmark and exhibits robust out-of-distribution generalization, substantially outperforming strong MLLM baselines and SFT approaches. A key contribution of this work is also the introduction of the XHSPost benchmark, a novel resource for evaluating TIMT in realistic social media contexts, which we hope will spur further research in this domain. Key analytical insights reveal the benefits of zero-start RL, the crucial synergy of our multi-task design, and the efficacy of a mixed-metric reward strategy for translation. Our work underscores the potential of multi-task RL for complex multimodal applications and offers valuable guidelines for developing more sophisticated RL-driven MLLMs for TIMT and related domains.

\noindent{\textbf{Limitations and Future Work.}} Although MT$^{3}$ demonstrates strong performance on the TIMT task across multiple benchmarks, there remains room for improvement in its multilingual capabilities and user interaction. In future work, we aim to leverage the robust text-grounding and visual understanding abilities of MLLMs to extend this approach to cross-lingual and multilingual VQA and reasoning tasks, while further enhancing MLLMs' alignment with user preferences.


\clearpage
\bibliographystyle{unsrtnat}
\bibliography{custom}


\clearpage
\appendix

\section{Evaluation Details}
\label{app:eval}
When evaluating MLLM's performance on the test set,  we deployed open-source models locally using frameworks like vLLM\footnote{\url{https://github.com/vllm-project/vllm}}  or lmdeploy\footnote{\url{https://github.com/InternLM/lmdeploy}} implementations.  We use the greedy decoding strategy for all systems. The maximum generation length was capped at 4096 tokens.
The prompt for evaluating MLLMs is structured as follows:
\begin{tcolorbox}[
    colframe=teal!70!black, 
    colback=teal!10!white, 
    coltitle=white, 
    fonttitle=\bfseries, 
    title=Evaluating MLLMs Prompt, 
    sharp corners, 
    boxrule=0.5mm, 
]
System: You are a helpful translation assistant. The user provides an image containing \textit{\{source\_language\}} text and asks for the corresponding \textit{\{target\_language\}} translation. \\
User: \textit{\{image\}} Translate the text in the image from \textit{\{source\_language\}} into \textit{\{target\_language\}} following the natural reading order without any explanation.
\label{prompt:evaluate_mllm}
\end{tcolorbox}

\section{SFT Prompts}
\label{app:sft_prompts}
The prompt for SFT (TIMT) is structured as follows:

\begin{tcolorbox}[
    colframe=teal!70!black, 
    colback=teal!10!white, 
    coltitle=white, 
    fonttitle=\bfseries, 
    title=SFT (TIMT) Prompt, 
    sharp corners, 
    boxrule=0.5mm, 
]
System: You are a helpful translation assistant. The user provides an image containing \textit{\{source\_language\}} text and asks for the corresponding \textit{\{target\_language\}} translation. Then, the assistant provides the user with the final translation in reading order, separating text from different positions (boxes) with a line break. The final translation must be enclosed within <translate> </translate> tags. The format must be as follows: <translate>final translation here</translate> \\
User: \textit{\{image\}} Translate all the text in this image into \textit{\{target\_language\}} following the natural reading order. \\
Assistant: <translate> \textit{\{target\_text\}} </translate>
\label{prompt:wo_recognize_think}
\end{tcolorbox}

The prompt for SFT (OCR + TIMT) is structured as follows:

\begin{tcolorbox}[
    colframe=teal!70!black, 
    colback=teal!10!white, 
    coltitle=white, 
    fonttitle=\bfseries, 
    title=SFT (OCR + TIMT) Prompt, 
    sharp corners, 
    boxrule=0.5mm, 
]
System: You are a helpful translation assistant. The user provides an image containing \textit{\{source\_language\}} text and asks for the corresponding \textit{\{target\_language\}} translation. First, the assistant recognizes all the text in the image following the natural reading order, separating text from different positions (boxes) with a line break. Then, the assistant provides the user with the final translation in reading order, separating text from different positions (boxes) with a line break. The recognized text and final translation must be enclosed within <recognize> </recognize> and <translate> </translate> tags, respectively. The format must be as follows: <recognize>recognized text here</recognize> <translate>final translation here</translate> \\
User: \textit{\{image\}} Translate all the text in this image into \textit{\{target\_language\}} following the natural reading order. \\
Assistant: <recognize> \textit{\{source\_text\}} </recognize><translate> \textit{\{target\_text\}} </translate>
\end{tcolorbox}

\section{XHSPost Benchmark Construction Details}
\label{app:xhspost_construction}
Social media platforms like TikTok, Instagram, and Xiaohongshu (also known as Red Note) have significantly deepened intercultural communication worldwide. However, there is a notable absence of evaluation benchmarks specifically designed for TIMT in social media contexts. 

\begin{figure}[h]
\centering
\includegraphics[width=0.92\textwidth]{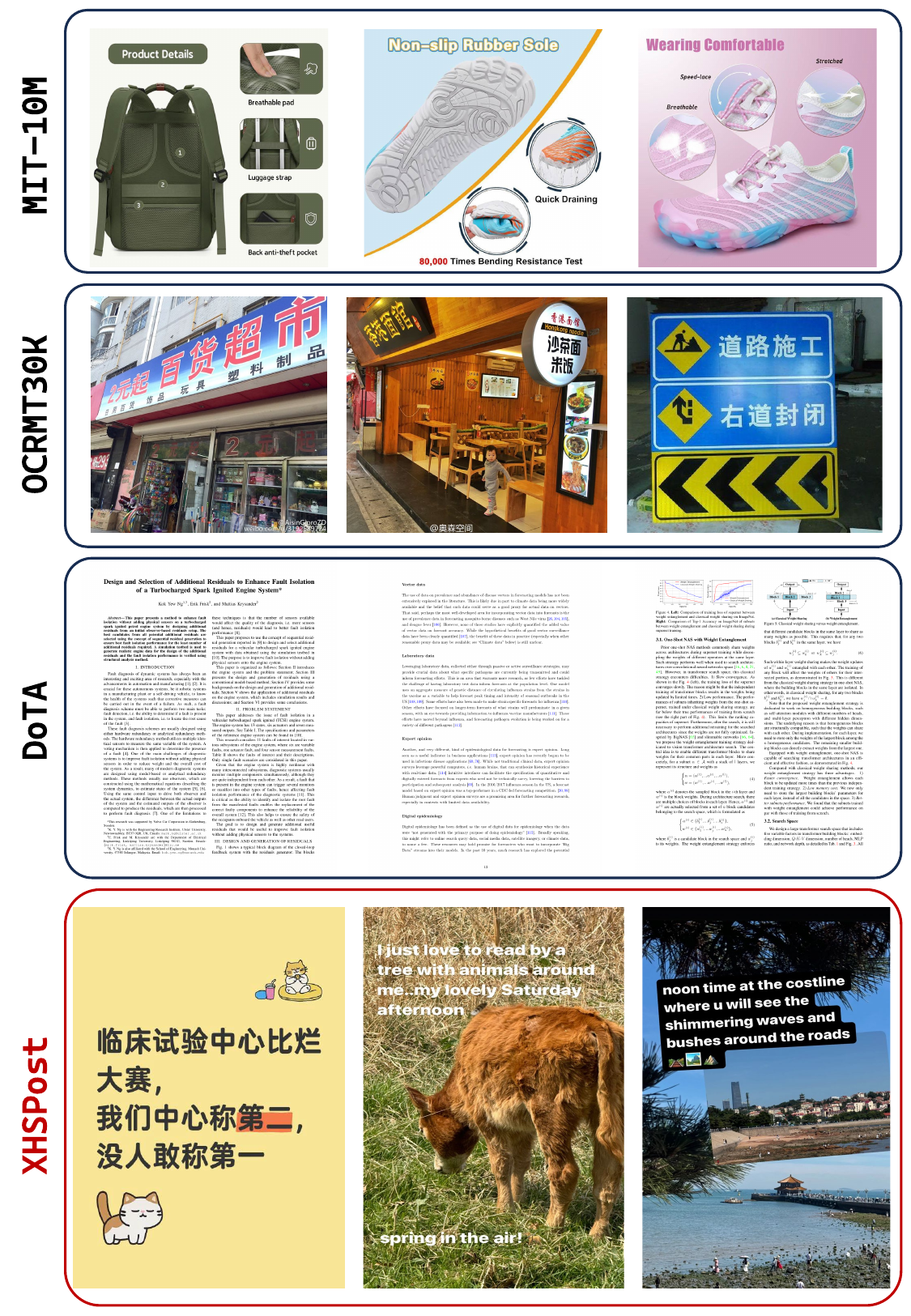}
\caption{
Comparison of XHSPost with other TIMT datasets.
}
\label{fig:dataset_case}
\end{figure}

The XHSPost benchmark was constructed to evaluate TIMT in real-world social media contexts, specifically using posts from Xiaohongshu, a globally-oriented platform. The construction process involved several key steps:

\begin{enumerate}
    \item \textbf{Data Collection}: We targeted posts containing either entirely English or entirely Chinese text. Initial post candidates were gathered by querying keywords relevant to common social media usage patterns, such as 'PLOG' (Photo Log), '\chinese{中文}Plog' (Chinese Plog), and 'TikTok Refugee'.
    \item \textbf{Content Appropriateness Filtering}: Following \citet{li2024mit10m}, all collected images underwent an NSFW detection process using an established tool~\citep{nsfw}. Any images flagged as potentially inappropriate were discarded.
    \item \textbf{Text Recognition and Viability Check}: We utilized GPT-4o (OpenAI's \texttt{gpt-4o-2024-11-20}) for detailed text recognition from the images. Posts where no precise textual information could be reliably extracted were excluded from further processing.
    \item \textbf{Personal Data and Sensitive Content Removal}: To mitigate the risk of exposing personal data, images where the OCR-extracted text contained sensitive information (e.g., email addresses, phone numbers) were removed. Additionally, images containing NSFW characters in the recognized text were also excluded to further ensure the dataset's appropriateness.
    \item \textbf{Final Dataset Curation}: This multi-stage filtering process yielded a curated set of 106 English posts and 109 Chinese posts.
    \item \textbf{Translation and Post-Editing}: For each selected post, the OCR-recognized text was translated by GPT-4o, using the corresponding image as contextual input to aid translation accuracy and appropriateness. Crucially, all machine-generated translations underwent manual post-editing by human reviewers to ensure high linguistic quality, fluency, and fidelity to the source text in its visual context. This step was vital for creating reliable ground-truth translations.
\end{enumerate}
This process resulted in the creation of two parallel datasets: XHSPost (EN-ZH) and XHSPost (ZH-EN), designed to support research in social media TIMT.
Illustrative examples can be found in Figure~\ref{fig:xhs_case} and \ref{fig:dataset_case}. With this work, we have taken an initial step in addressing TIMT for social media scenarios. We anticipate that this benchmark will encourage further research in this area, ultimately contributing to enhanced cross-cultural communication. The benchmark will be released upon acceptance.

\section{Training Details}
\label{app:train_details}
We employ a constant learning rate of 5e-7 and set the sampling temperature to 1.0.
The maximum generation length for responses is capped at
4096 tokens.
We set the KL penalty coefficient $\beta$
to 0.01 and set the PPO clipping range $\epsilon$ to 0.2.
The model is trained for 1 epoch on 32 NVIDIA
H800 80G GPUs for about 15 hours.

\section{OOD BLEU Score}
\label{app:ood_bleu}
Due to the scope limits of the paper, we provide the BLEU score for these OOD evaluations in Table~\ref{tab:ood_bleu}.
\input{table/ood_bleu}

\section{Multi-Task Ablation Prompts}
\label{app:multi_task_prompt}
The prompts used for multi-task ablation are as follows:
\begin{tcolorbox}[
    colframe=teal!70!black, 
    colback=teal!10!white, 
    coltitle=white, 
    fonttitle=\bfseries, 
    title=w/o Reasoning, 
    sharp corners, 
    boxrule=0.5mm, 
]
System: You are a helpful translation assistant. The user provides an image containing \textit{\{source\_language\}} text and asks for the corresponding \textit{\{target\_language\}} translation. First, the assistant recognizes all the text in the image following the natural reading order, separating text from different positions (boxes) with a line break. Then, the assistant provides the user with the final translation in reading order, separating text from different positions (boxes) with a line break. The recognized text and final translation must be enclosed within <recognize> </recognize> and <translate> </translate> tags, respectively. The format must be as follows: <recognize>recognized text here</recognize> <translate>final translation here</translate> \\
User: \textit{\{image\}} Translate all the text in this image into \textit{\{target\_language\}} following the natural reading order.
\end{tcolorbox}

\begin{tcolorbox}[
    colframe=teal!70!black, 
    colback=teal!10!white, 
    coltitle=white, 
    fonttitle=\bfseries, 
    title=w/o Recognition, 
    sharp corners, 
    boxrule=0.5mm, 
]
System: You are a helpful translation assistant. The user provides an image containing \textit{\{source\_language\}} text and asks for the corresponding \textit{\{target\_language\}} translation. First, the assistant carefully analyzes the recognized text and the visual elements in the image, considering layout, objects, color schemes, spatial relationships, and other contextual clues that may influence meaning. This integrated understanding ensures the translation is accurate, coherent, and appropriate to the visual setting. After thorough reasoning based on both textual content and visual context, the assistant provides the user with the final translation in reading order. The reasoning process and final translation are enclosed within <think> </think> and <translate> </translate> tags, respectively. The format must be as follows: <think> reasoning process here </think><translate> final translation here </translate> \\
User: \textit{\{image\}} Translate all the text in this image into \textit{\{target\_language\}} following the natural reading order.
\end{tcolorbox}

\begin{tcolorbox}[
    colframe=teal!70!black, 
    colback=teal!10!white, 
    coltitle=white, 
    fonttitle=\bfseries, 
    title=Only TIMT, 
    sharp corners, 
    boxrule=0.5mm, 
]
System: You are a helpful translation assistant. The user provides an image containing \textit{\{source\_language\}} text and asks for the corresponding \textit{\{target\_language\}} translation. Then, the assistant provides the user with the final translation in reading order, separating text from different positions (boxes) with a line break. The final translation must be enclosed within <translate> </translate> tags. The format must be as follows: <translate>final translation here</translate> \\
User: \textit{\{image\}} Translate all the text in this image into \textit{\{target\_language\}} following the natural reading order. 
\end{tcolorbox}

\section{Detailed Analysis of XHSPost Example}
\label{app:xhs_case_analysis}
As illustrated in Figure~\ref{fig:xhs_case}, MT$^{3}$ correctly translate culturally nuanced terms (like the TV show \textit{'Friends'} to its official Chinese title \textit{\chinese{'老友记'}} by inferentially linking the visual cue on a phone screen to its broader context, unlike models that produce literal translations such as \textit{\chinese{'朋友'}} (friendship)) and accurately interpret ambiguous numerical information in advertisements (discerning price/location from '800' instead of '800 square meters') stems from this integrated approach.

\section{Cold-Start Initialization}
\label{app:initial}
In this work, we utilize QVQ \citep{qvq-72b-preview} alongside the ground truth TIMT dataset to construct the cold-start dataset.
During preliminary experiments, we observed that as a reasoning model, QVQ exhibits limited instruction-following capabilities, rendering it unsuitable for directly applying the MT$^{3}$ prompt to generate cold-start data.
Instead, we provide QVQ with images and corresponding ground truth source texts from the TIMT dataset, and request only the reasoning process and final translation.
We keep most parts of the prompt remain consistent with the MT$^{3}$ prompt:
:

\begin{tcolorbox}[
    colframe=teal!70!black, 
    colback=teal!10!white, 
    coltitle=white, 
    fonttitle=\bfseries, 
    title=Cold-Start Data Construction Prompt, 
    sharp corners, 
    boxrule=0.5mm, 
]
System: You are a helpful translation assistant. The user provides an image containing \textit{\{source\_language\}} text and asks for the corresponding \textit{\{target\_language\}} translation. First, the assistant carefully analyzes the recognized text together with the visual elements in the image, taking into account the layout, objects, color schemes, spatial relationships, and other contextual clues that may influence meaning. This integrated understanding ensures that the translation is accurate, coherent, and appropriate to the visual setting. After thorough reasoning based on both textual content and visual context, the assistant provides the user with the final translation in reading order. You should output your final \textit{\{target\_language\}} translation as: Final Translation: Your final translation. You should think step-by-step.\\
User: \textit{\{image\}} Translate this text from \textit{\{source\_language\}} to \textit{\{target\_language\}}. \textit{\{source\_language\}}: \textit{\{source\_text\}}
\end{tcolorbox}

For quality control, we extract the final translation of each sample using the regular expression, compute the BLEU score of each translation against the ground truth target translation, and select the top 5K samples with the highest BLEU scores for both EN-ZH and ZH-EN directions.

Subsequently, we merge the ground truth source texts with the QVQ-generated reasoning processes and translations, following the MT$^{3}$ prompt format to construct the final cold-start dataset.
This dataset is further used for the cold-start SFT initialization of MT$^{3}$.

\section{Case Study: Initialization Strategy Impact on Reasoning}
\label{app:case}
\begin{figure}[ht]
\centering
\includegraphics[width=1.0\textwidth]{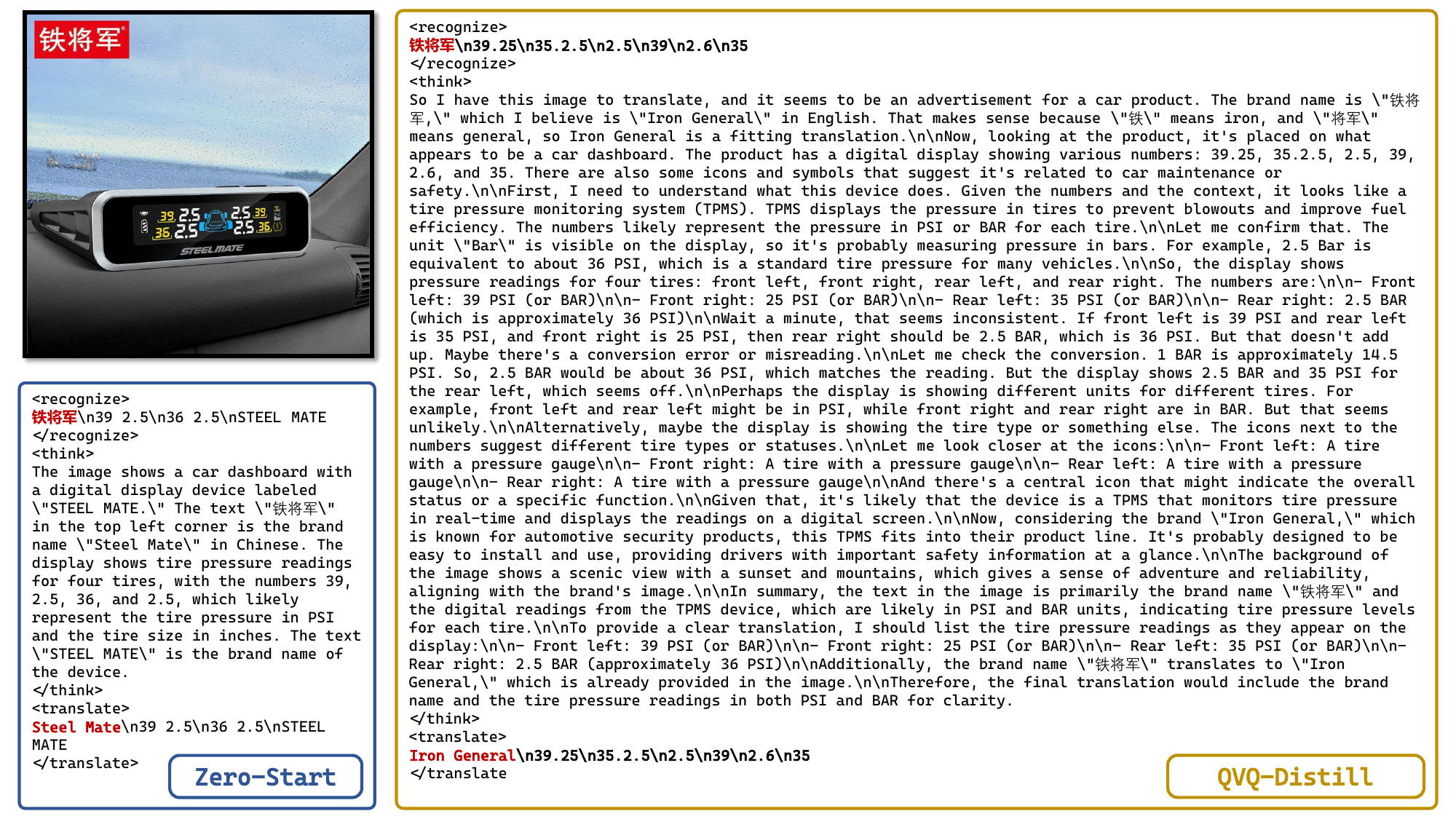}
\caption{
Case Study. The model initialized with QVQ-Distill tends to generate significantly longer reasoning steps. In contrast, the Zero-start model, after RL training, often develops a more straightforward and less redundant reasoning path.
}
\label{fig:mt3_case}
\end{figure}

We provide a qualitative case study in Figure~\ref{fig:mt3_case} to illustrate the differences in reasoning patterns that emerge from RL training with different initialization strategies: MT$^{3}$-7B-Zero (\textit{Zero-start}) versus MT$^{3}$-7B-QVQ-Distill (\textit{QVQ-Distill}). As observed in the examples, the model initialized with QVQ-Distill tends to generate significantly longer reasoning steps within the \texttt{<think>} tags. These longer reasoning chains often exhibit self-reflection patterns that appear to be cloned or heavily influenced by the QVQ teacher model. In contrast, the Zero-start model, after RL training, often develops a more straightforward and less redundant reasoning path. This more concise reasoning appears to be directly focused on the core elements necessary for the TIMT task, potentially contributing to its superior performance and more efficient processing.

\section{Difficulty Levels for Curriculum Learning}
\label{app:difficulty} 
The curriculum learning experiments (Section~\ref{sec:analysis_curriculum}) utilize difficulty categorizations for data instances from the MIT-10M dataset~\citep{li2024mit10m}. During the construction of the MIT-10M dataset, samples were classified into three difficulty levels based on heuristics considering the token length of the in-image text and the number of detected bounding boxes for text regions. These categories are defined as follows:

\begin{itemize}
    \item \textbf{Easy:} Instances where the number of bounding boxes is less than or equal to 2, and the token length of the text is less than or equal to 16. These samples generally contain fewer textual elements and shorter texts, resulting in a relatively straightforward translation task.
    \item \textbf{Hard:} Instances where the number of bounding boxes is greater than 5, or the token length of the text is greater than 25. These samples typically feature more textual elements or longer, more complex texts, placing higher demands on the model's attention mechanism and processing capabilities.
    \item \textbf{Medium:} All other instances that do not fall into the Easy or Hard categories. These samples exhibit a wider spread in terms of the number of bounding boxes and text token lengths, representing more realistic and diverse image translation scenarios. This category is particularly useful for testing the generalization ability of multimodal translation models.
\end{itemize}



\clearpage

\end{document}

%% file: table/main.tex
\begin{table*}[t]
    \centering
    \small
    \caption{
 In-domain (IND) performance comparison on the MIT-10M benchmark (ZH-EN and EN-ZH). Metrics reported are BLEU, chrF++, and METEOR, along with their average (Avg.). MT$^{3}$-7B-Zero (RL) is compared against open-source systems and supervised fine-tuned variants.
    }
    \label{tab:main_mt_results}
    \resizebox{0.95\textwidth}{!}{%
    \setlength{\tabcolsep}{4pt}
    \begin{tabular}{l*{9}{c}} 
        \toprule
        \multirow{2.5}{*}{\sc Model} & \multicolumn{4}{c}{\sc MIT-10M (ZH-EN)} & & \multicolumn{4}{c}{\sc MIT-10M (EN-ZH)} \\ 
        \cmidrule(lr){2-5} \cmidrule(lr){7-10} 
         & BLEU & chrF++ & METEOR & Avg. & & BLEU & chrF++ & METEOR & Avg. \\ 
        \midrule

        \multicolumn{10}{c}{\textit{\textbf{Zero-shot Systems}}} \\ 
        \multicolumn{10}{@{}l}{\textcolor{lightgray}{\textit{Cascade System}}} \\
        EasyOCR + NLLB-3.3B   & 2.54 & 10.71 & 7.72 & 6.99 & & 11.93 & 12.48 & 10.84 & 11.75 \\ 

        \multicolumn{10}{@{}l}{\textcolor{lightgray}{\textit{Advanced MLLMs}}} \\
        InternVL2.5-8B   & 9.98 & 26.59 & 24.99 & 20.52 & & 31.74 & 38.21 & 38.49 & 36.15 \\
        InternVL2.5-26B   & 13.29 & 27.82 & 24.60 & 21.90 & & 33.22 & 42.43 & 46.55 & 40.73 \\
        InternVL2.5-38B   & 10.77 & 26.00 & 24.49 & 20.42 & & 38.61 & 48.87 & 44.04 & 43.84 \\
        InternVL2.5-78B   & 12.50 & 26.85 & 27.11 & 22.15 & & 36.97 & 48.96 & 46.19 & 44.04 \\
        Qwen2.5-VL-7B    & 15.59 & 31.34 & 33.44 & 26.79 & & 31.66 & 44.20 & 48.41 & 41.42 \\
        Qwen2.5-VL-72B   & 11.33 & 32.82 & 41.26 & 28.47 & & 35.55 & 50.78 & 52.76 & 46.36 \\

        \midrule

        \multicolumn{10}{c}{\textit{\textbf{Fine-tuned VLMs}}} \\ 
        Qwen2.5-VL-7B (SFT, TIMT)    & 18.66 & 40.74 & 43.20 & 34.20 & & 43.33 & 48.62 & 55.71 & 49.22 \\
        Qwen2.5-VL-7B (SFT, OCR + TIMT)    & 22.57 & 42.39 & 41.80 & 35.59 & & 48.69 & 54.66 & 59.60 & 54.32 \\
        MT$^{3}$-7B-Zero (RL) & 20.31 & 44.42 & 43.99 & 36.24 & & 54.49 & 61.77 & 68.23 & 61.50 \\

        \bottomrule
    \end{tabular}
    }
\end{table*}

%% file: table/ood_chrf_meteor.tex
\begin{table}[t]
    \centering
    \caption{
Out-of-distribution (OOD) generalization performance using chrF++ and METEOR. MT$^{3}$-7B-Zero (RL) is evaluated against strong and same-size MLLM baselines on unseen language pairs from MIT-10M (EN-DE, ZH-FR, DE-FR) and different OOD datasets (OCRMT30K, DoTA, XHSPost). \textbf{Bold} indicates the best, \underline{underline} indicates the second best.
    }
    \label{tab:ood_chrf_meteor}
    \resizebox{0.98\columnwidth}{!}{%
    \setlength{\tabcolsep}{2pt}
    \begin{tabular}{@{}lcccccccccccccccc@{}}
        \toprule
        \multirow{3}{*}{\sc Model} 
        & \multicolumn{6}{c}{\sc OOD (Language Pairs)}  
        & \multicolumn{8}{c}{\sc OOD (Datasets)} \\
        \cmidrule(lr){2-7} 
        \cmidrule(lr){8-15}
        & \multicolumn{2}{c}{\sc MIT-10M (EN-DE)} 
        & \multicolumn{2}{c}{\sc MIT-10M (ZH-FR)} 
        & \multicolumn{2}{c}{\sc MIT-10M (DE-FR)} 
        & \multicolumn{2}{c}{\sc OCRMT30K (ZH-EN)} 
        & \multicolumn{2}{c}{\sc DoTA (EN-ZH)} 
        & \multicolumn{2}{c}{\sc XHSPost (EN-ZH)} 
        & \multicolumn{2}{c}{\sc XHSPost (ZH-EN)} \\
        \cmidrule(lr){2-3} \cmidrule(lr){4-5} \cmidrule(lr){6-7} 
        \cmidrule(lr){8-9} \cmidrule(lr){10-11} \cmidrule(lr){12-13} \cmidrule(lr){14-15}
        & chrF++ & METEOR & chrF++ & METEOR & chrF++ & METEOR 
        & chrF++ & METEOR & chrF++ & METEOR & chrF++ & METEOR & chrF++ & METEOR \\
        \midrule
        \multicolumn{15}{@{}l}{\textcolor{lightgray}{\textit{Strong Baseline}}} \\
        Qwen2.5-VL-72B & 48.37 & 40.60 & \underline{27.29} & 30.13 & \underline{48.82} & \underline{45.06} 
          & \textbf{39.73} & \textbf{42.65} & \textbf{37.65} & \textbf{50.75} 
          & \textbf{42.97} & \textbf{61.28} & \textbf{55.10} & \textbf{54.07} \\
        InternVL2.5-78B & \underline{50.54} & \underline{43.11} & 18.25 & 19.59 & 27.98 & 25.45 
          & 31.57 & 32.81 & 32.88 & 45.03 
          & 37.61 & 57.60 & 53.14 & 51.18 \\
        \multicolumn{15}{@{}l}{\textcolor{lightgray}{\textit{Same-size Baseline}}} \\
        Qwen2.5-VL-7B & 48.12 & 40.65 & 27.19 & \underline{30.59} & 33.60 & 31.99 
          & 23.99 & 24.90 & 32.62 & 44.16 
          & 36.79 & 55.16 & 49.01 & 47.05 \\
        InternVL2.5-8B & 42.42 & 36.67 & 19.32 & 14.71 & 23.17 & 20.41 
          & 21.95 & 21.12 & 24.75 & 34.89 
          & 28.32 & 46.20 & 42.19 & 38.93 \\
        \midrule
        MT$^{3}$-7B-Zero (RL) & \textbf{56.14} & \textbf{52.11} & \textbf{35.67} & \textbf{38.53} & \textbf{53.88} & \textbf{54.99} 
          & \underline{37.42} & \underline{38.17} & \underline{34.87} & \underline{46.60} 
          & \underline{39.95} & \underline{58.58} & \underline{53.26} & \underline{52.04} \\
        \bottomrule
    \end{tabular}
    }
\vspace{-3mm}
\end{table}

%% file: table/task_ablation.tex
\begin{table*}[h]
    \centering
    \small
    \caption{
 Multi-task ablation study for MT$^{3}$ framework. Performance impact of removing Recognition, Reasoning, or both, compared to the full task setup. $\Delta$ denotes the average metric score difference compared to the \textit{Full Tasks} setting.
    }
    \label{tab:ablation_results}
    \resizebox{\textwidth}{!}{%
    \setlength{\tabcolsep}{4pt}
    \begin{tabular}{lccc*{4}{c}c*{4}{c}c*{4}{c}c}
        \toprule
        \multirow{2.5}{*}{\sc Ablation} & \multicolumn{3}{c}{\sc Tasks} 
        & \multicolumn{4}{c}{\sc MIT-10M (ZH-EN)} & \multirow{2.5}{*}{\sc $\Delta$}
        & \multicolumn{4}{c}{\sc MIT-10M (EN-ZH)} & \multirow{2.5}{*}{\sc $\Delta$}
        & \multicolumn{4}{c}{\sc XHSPost (ZH-EN)} & \multirow{2.5}{*}{\sc $\Delta$} \\
        \cmidrule(lr){2-4} \cmidrule(lr){5-8} \cmidrule(lr){10-13} \cmidrule(lr){15-18}
         & Recognition & Reasoning & Translation 
         & BLEU & chrF++ & METEOR & Avg. & 
         & BLEU & chrF++ & METEOR & Avg. & 
         & BLEU & chrF++ & METEOR & Avg. & \\
        \midrule
        Full Tasks & \cmark & \cmark & \cmark 
            & 20.31 & 44.42 & 43.99 & 36.24 & ---
            & 54.49 & 61.77 & 68.23 & 61.50 & ---
            & 26.94 & 53.26 & 52.04 & 44.08 & --- \\
        w/o Reasoning & \xmark & \cmark & \cmark 
            & 22.21 & 45.06 & 44.63 & 37.30 & +1.06
            & 52.12 & 61.34 & 64.26 & 59.24 & -2.26
            & 25.01 & 51.50 & 49.73 & 42.08 & -2.00 \\
        w/o Recognition & \cmark & \xmark & \cmark 
            & 16.99 & 40.08 & 42.65 & 33.24 & -3.00
            & 47.61 & 57.78 & 65.11 & 56.83 & -4.67
            & 25.06 & 52.01 & 50.67 & 42.58 & -1.50 \\
        Only TIMT & \xmark & \xmark & \cmark 
            & 13.43 & 38.47 & 42.21 & 31.37 & -4.87
            & 44.04 & 55.87 & 61.77 & 53.89 & -7.61
            & 24.21 & 51.32 & 49.54 & 41.69 & -2.39 \\
        \bottomrule
    \end{tabular}
    }
\end{table*}

%% file: table/think_ablation.tex
\begin{table}[t]
    \centering
    \caption{
Impact of initialization strategy on performance. Comparison of MT$^{3}$-7B-Zero (RL from scratch) and MT$^{3}$-7B-QVQ-Distill (SFT on distilled data then RL) on in-domain (MIT-10M) and out-of-distribution (OCRMT30K, DoTA, XHSPost) benchmarks. 
    }
    \label{tab:think_ablation}
    \resizebox{0.95\columnwidth}{!}{%
    \setlength{\tabcolsep}{2pt}
    \begin{tabular}{@{}lccccccccccccccccccc@{}}
        \toprule
        \multirow{3}{*}{\sc Model} 
        & \multicolumn{6}{c}{\sc In-Domain}  
        & \multicolumn{9}{c}{\sc Out-of-Domain} \\
        \cmidrule(lr){2-7} \cmidrule(lr){8-19}
        & \multicolumn{3}{c}{\sc MIT-10M (ZH-EN)} 
        & \multicolumn{3}{c}{\sc MIT-10M (EN-ZH)} 
        & \multicolumn{3}{c}{\sc OCRMT30K (ZH-EN)} 
        & \multicolumn{3}{c}{\sc DoTA (EN-ZH)} 
        & \multicolumn{3}{c}{\sc XHSPost (EN-ZH)} 
        & \multicolumn{3}{c}{\sc XHSPost (ZH-EN)} \\
        \cmidrule(lr){2-4} \cmidrule(lr){5-7} 
        \cmidrule(lr){8-10} \cmidrule(lr){11-13} \cmidrule(lr){14-16} \cmidrule(lr){17-19}
        & BLEU & chrF++ & METEOR 
        & BLEU & chrF++ & METEOR 
        & BLEU & chrF++ & METEOR 
        & BLEU & chrF++ & METEOR 
        & BLEU & chrF++ & METEOR 
        & BLEU & chrF++ & METEOR \\
        \midrule
        \multicolumn{19}{@{}l}{\textcolor{lightgray}{\textit{w/ "Cold Start"}}} \\
        MT$^{3}$-7B-QVQ-Distill & 
            15.37 & 36.36 & 40.20 & 
            42.21 & 53.55 & 58.37 & 
            12.39 & 34.68 & 35.43 & 
            29.23 & 25.60 & 37.08 & 
            48.20 & 39.61 & 58.41 & 
            24.61 & 49.74 & 48.76 \\
        \multicolumn{19}{@{}l}{\textcolor{lightgray}{\textit{w/o "Cold Start"}}} \\
        MT$^{3}$-7B-Zero & 
            20.31 & 44.42 & 43.99 & 
            54.49 & 61.77 & 68.23 & 
            14.21 & 37.42 & 38.17 & 
            40.46 & 34.87 & 46.60 & 
            49.41 & 39.95 & 58.58 & 
            26.94 & 53.26 & 52.04 \\
        \bottomrule
    \end{tabular}
    }
    \vspace{-3mm}
\end{table}

%% file: table/ood_bleu.tex
\begin{table}[h]
    \centering
    \caption{
    OOD BLEU score performance. Complement chrF++ and METEOR results in Table~\ref{tab:ood_chrf_meteor}.
    }
    \label{tab:ood_bleu}
    \resizebox{\columnwidth}{!}{%
    \setlength{\tabcolsep}{4pt}
    \begin{tabular}{@{}lcccccccc@{}}
        \toprule
        \multirow{3}{*}{\sc Model} 
        & \multicolumn{3}{c}{\sc OOD (Language Pairs)}  
        & \multicolumn{5}{c}{\sc OOD (Datasets)}\\
        \cmidrule(lr){2-4} \cmidrule(lr){5-9}
        & \sc MIT-10M (EN-DE) 
        & \sc MIT-10M (ZH-FR) 
        & \sc MIT-10M (DE-FR) 
        & \sc OCRMT30K (ZH-EN) 
        & \sc DoTA (EN-ZH) 
        & \sc XHSPost (EN-ZH) 
        & \sc XHSPost (ZH-EN) \\
        \midrule
        \multicolumn{9}{@{}l}{\textcolor{lightgray}{\textit{Strong Baseline}}} \\
        Qwen2.5-VL-72B    & 20.49 & 9.79  & 18.78 & 15.14 & 45.43 & 52.97 & 29.52 \\
        InternVL2.5-78B   & 21.23 & 5.22  & 11.16 & 11.46 & 38.28 & 45.61 & 28.87 \\
        \multicolumn{9}{@{}l}{\textcolor{lightgray}{\textit{Same-size Baseline}}} \\
        Qwen2.5-VL-7B     & 22.85 & 12.09 & 14.99 & 8.93  & 36.90 & 44.96 & 24.49 \\
        InternVL2.5-8B    & 16.76 & 6.68  & 8.49  & 7.10  & 27.25 & 35.74 & 19.21 \\
        \midrule
        MT$^{3}$-7B-Zero & 29.50 & 15.97 & 26.92 & 14.21 & 40.46 & 49.41 & 26.94 \\
        \bottomrule
    \end{tabular}
    }
\end{table}